\documentclass[runningheads]{llncs}

\usepackage{graphicx}
\usepackage{amsmath,amssymb} 
\usepackage{color}
\usepackage{multirow}       
\usepackage[linesnumbered,ruled,vlined]{algorithm2e} 
\usepackage{array}
\usepackage[numbers,sort]{natbib}
\graphicspath{ {figures/} }      
\DeclareGraphicsExtensions{.pdf} 

\begin{document}

\newcolumntype{C}{>{\centering\arraybackslash}p{4.4em}}

\title{NetAdapt: Platform-Aware Neural Network Adaptation for Mobile Applications}

\titlerunning{NetAdapt}
%

\author{Tien-Ju Yang\inst{1}\thanks{This work
was done while Tien-Ju Yang was an intern at Google.}\orcidID{0000-0003-4728-0321} \and Andrew Howard\inst{2} \and Bo Chen\inst{2} \and \\ Xiao Zhang\inst{2} \and Alec Go\inst{2} \and Mark Sandler\inst{2} \and Vivienne Sze\inst{1} \and Hartwig Adam\inst{2}}

%
\authorrunning{T.-J. Yang et al.}
%

\institute{Massachusetts Institute of Technology\\
\and
Google Inc.\\
\{tjy,sze\}@mit.edu, \{howarda,bochen,andypassion,ago,sandler,hadam\}@google.com}
\maketitle              
%

\begin{abstract}
This work proposes an algorithm, called NetAdapt, that \emph{automatically adapts} a pre-trained deep neural network to a mobile platform given a resource budget. While many existing algorithms simplify networks based on the number of MACs or weights, optimizing those indirect metrics may not necessarily reduce the direct metrics, such as latency and energy consumption. To solve this problem, NetAdapt incorporates direct metrics into its adaptation algorithm. These direct metrics are evaluated using \emph{empirical measurements}, so that detailed knowledge of the platform and toolchain is not required. NetAdapt automatically and progressively simplifies a pre-trained network until the resource budget is met while maximizing the accuracy. Experiment results show that NetAdapt achieves better accuracy versus latency trade-offs on both mobile CPU and mobile GPU, compared with the state-of-the-art automated network simplification algorithms. For image classification on the ImageNet dataset, NetAdapt achieves up to a 1.7$\times$ speedup in \emph{measured inference latency} with equal or higher accuracy on MobileNets (V1\&V2).
\end{abstract}


\section{Introduction}
\label{sec:introduction}

Deep neural networks (DNNs or networks) have become an indispensable component of artificial intelligence, delivering near or super-human accuracy on common vision tasks such as image classification and object detection. However, DNN-based AI applications are typically too computationally intensive to be deployed on resource-constrained platforms, such as mobile phones. This hinders the enrichment of a large set of user experiences.

A significant amount of recent work on DNN design has focused on improving the efficiency of networks. However, the majority of works are based on optimizing the ``indirect metrics'', such as the number of multiply-accumulate operations (MACs) or the number of weights, as proxies for the resource consumption of a network. Although these indirect metrics are convenient to compute and integrate into the optimization framework, they may not be good approximations to the ``direct metrics'' that matter for the real applications such as latency and energy consumption. The relationship between an indirect metric and the corresponding direct metric can be highly non-linear and platform-dependent as observed by ~\cite{cvpr2017-yang-energy-aware-pruning,isca2017-yu-scalpel,sysml2018-lai-not-all-ops-equal}. In this work, we will also demonstrate empirically that a network with a fewer number of MACs can be slower when actually running on mobile devices; specifically, we will show that a network of 19\% less MACs incurs 29\% longer latency in practice (see Table~\ref{table:mobilenet_cpu_benchmark_macs}).

There are two common approaches to designing efficient network architectures. The first is designing a single architecture with no regard to the underlying platform. It is hard for a single architecture to run optimally on all the platforms due to the different platform characteristics. For example, the fastest architecture on a desktop GPU may not be the fastest one on a mobile CPU with the same accuracy. Moreover, there is little guarantee that the architecture could meet the resource budget (e.g., latency) on all platforms of interest. The second approach is manually crafting architectures for a given target platform based on the platform's characteristics. However, this approach requires deep knowledge about the implementation details of the platform, including the toolchains, the configuration and the hardware architecture, which are generally unavailable given the proprietary nature of hardware and the high complexity of modern systems. Furthermore, manually designing a different architecture for each platform can be taxing for researchers and engineers.

\begin{figure}[!t]
    \centering
    \includegraphics[width=1.0\textwidth]{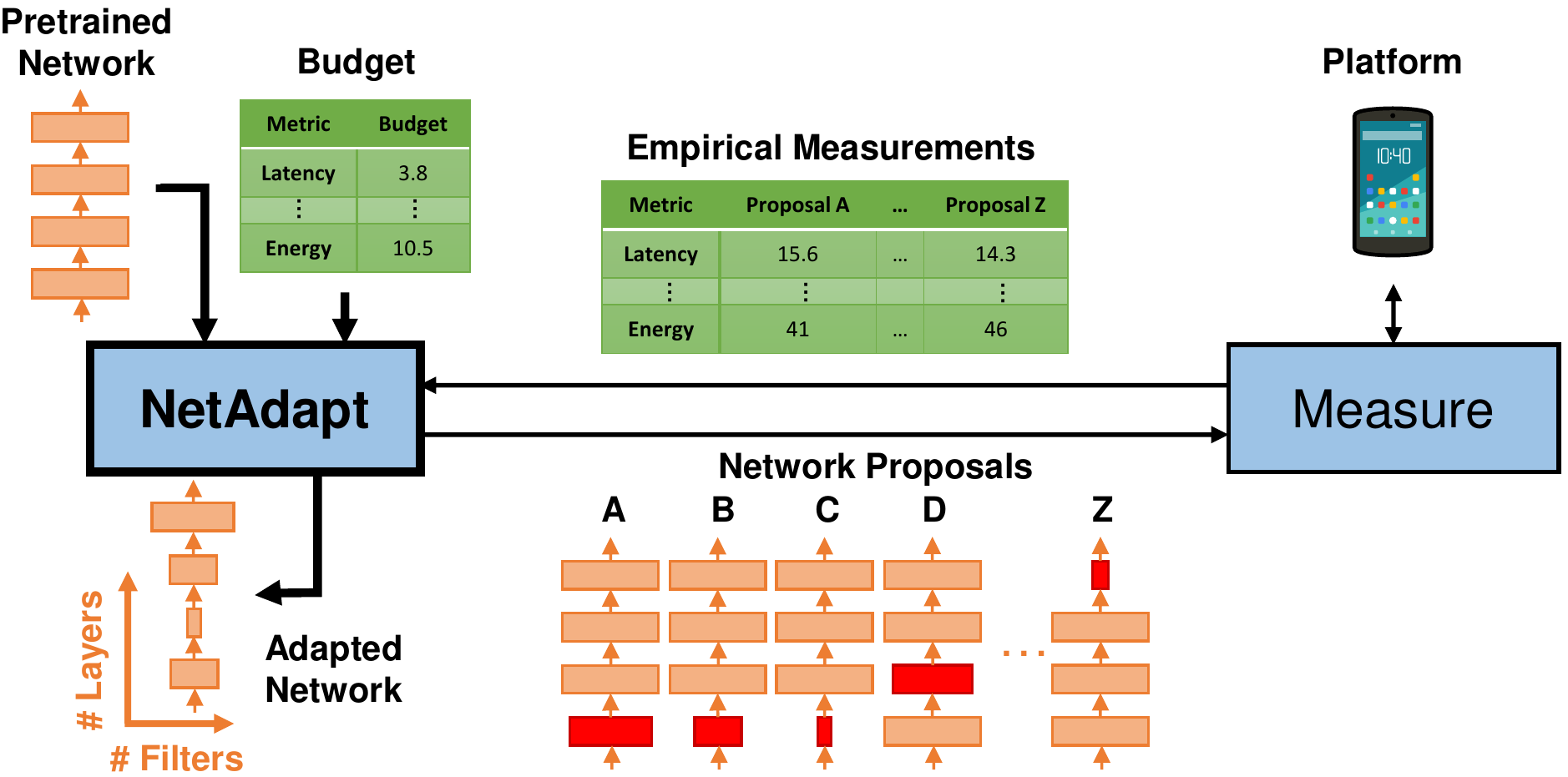}
    \caption{NetAdapt automatically adapts a pretrained network to a mobile platform given a resource budget. This algorithm is guided by the direct metrics for resource consumption. NetAdapt eliminates the requirement of platform-specific knowledge by using empirical measurements to evaluate the direct metrics. At each iteration, NetAdapt generates many network proposals and measures the proposals on the target platform. The measurements are used to guide NetAdapt to generate the next set of network proposals at the next iteration.}
    \label{fig:platform_aware_adaptation}
\end{figure}

In this work, we propose a platform-aware algorithm, called \textit{NetAdapt}, to address the aforementioned issues and facilitate platform-specific DNN deployment. NetAdapt (Fig.~\ref{fig:platform_aware_adaptation}) incorporates \textit{direct metrics} in the optimization loop, so it does not suffer from the discrepancy between the indirect and direct metrics. The direct metrics are evaluated by the empirical measurements taken from the target platform. This enables the algorithm to support any platform without detailed knowledge of the platform itself, although such knowledge could still be incorporated into the algorithm to further improve results. In this paper, we use latency as the running example of a direct metric and resource to target even though our algorithm is generalizable to other metrics or a combination of them (Sec.~\ref{subsec:results_on_macs}).

The network optimization of NetAdapt is carried out in an automatic way to gradually reduce the resource consumption of a pretrained network while maximizing the accuracy. The optimization runs iteratively until the resource budget is met. Through this design, NetAdapt can generate not only a network that meets the budget, but also a family of simplified networks with different trade-offs, which allows dynamic network selection and further study. Finally, instead of being a black box, NetAdapt is designed to be easy to interpret. For example, through studying the proposed network architectures and the corresponding empirical measurements, we can understand why a proposal is chosen and this sheds light on how to improve the platform and network design.

The main contributions of this paper are:
\begin{itemize}
     \item A framework that uses direct metrics when optimizing a pretrained network to meet a given resource budget.  Empirical measurements are used to evaluate the direct metrics such that no platform-specific knowledge is required.
     \item An automated constrained network optimization algorithm that maximizes accuracy while satisfying the constraints (i.e., the predefined resource budget). The algorithm outperforms the state-of-the-art automatic network simplification algorithms by up to 1.7$\times$ in terms of reduction in \emph{measured inference latency} while delivering equal or higher accuracy. Moreover, a family of simplified networks with different trade-offs will be generated to allow dynamic network selection and further study.
     \item Experiments that demonstrate the effectiveness of NetAdapt on different platforms and on real-time-class networks, such as the small MobileNetV1, which is more difficult to simplify than larger networks.
\end{itemize}

\section{Related Work}
\label{sec:related_work}

There is a large body of work that aims to simplify DNNs. We refer the readers to~\cite{pieee2017-sze-survey} for a comprehensive survey, and summarize the main approaches below.

The most related works are pruning-based methods. \cite{leoptimal,han2015learning,molchanov2016pruning} aim to remove individual redundant weights from DNNs. However, most  platforms cannot fully take advantage of unstructured sparse filters~\cite{isca2017-yu-scalpel}. Hu et al.~\cite{arxiv2016-hu-remove_zero_freq} and Srinivas et al.~\cite{srinivas2015data} focus on removing entire filters instead of individual weights. The drawback of these methods is the requirement of \emph{manually} choosing the compression rate for each layer. MorphNet~\cite{gordon2018morphnet} leverages the sparsifying regularizers to automatically determine the layerwise compression rate. ADC~\cite{he2018adc} uses reinforcement learning to learn a policy for choosing the compression rates. The crucial difference between all the aforementioned methods and ours is that they are not guided by the direct metrics, and thus may lead to sub-optimal performance, as we see in Sec.~\ref{subsec:results_on_macs}.

Energy-aware pruning~\cite{cvpr2017-yang-energy-aware-pruning} uses an energy model~\cite{asilomar2017-yang-estimate-energy} and incorporates the estimated energy numbers into the pruning algorithm. However, this requires designing models to estimate the direct metrics of each target platform, which requires detailed knowledge of the platform including its hardware architecture~\cite{jssc2016-chen}, and the network-to-array mapping used in the toolchain~\cite{isca2016-chen}. NetAdapt does not have this requirement since it can directly use empirical measurements.

DNNs can also be simplified by approaches that involve directly designing efficient network architectures, decomposition or quantization. MobileNets~\cite{howard2017mobilenets,cvpr2018-sandler-mobilenet-v2} and ShuffleNets~\cite{zhang2017shufflenet} provide efficient layer operations and reference architecture design. Layer-decomposition-based algorithms~\cite{kim2015compression,yang2015deep} exploit matrix decomposition to reduce the number of operations. Quantization~\cite{jacob2017quantization,hubara2016binarized,rastegari2016xnor} reduces the complexity by decreasing the computation accuracy. The proposed algorithm, NetAdapt, is complementary to these methods. For example, NetAdapt can adapt MobileNets to further push the frontier of efficient networks as shown in Sec.~\ref{sec:experiment_results} even though MobileNets are more compact and much harder to simplify than the other larger networks, such as VGG~\cite{iclr2015-simonyan-vgg}.

\section{Methodology: NetAdapt}
\label{sec:methodology}

We propose an algorithm, called NetAdapt, that will allow a user to automatically simplify a pretrained network to meet the resource budget of a platform while maximizing the accuracy. NetAdapt is guided by direct metrics for resource consumption, and the direct metrics are evaluated by using empirical measurements, thus eliminating the requirement of detailed platform-specific knowledge.

\subsection{Problem Formulation}
\label{subsec:problem_formulation}

NetAdapt aims to solve the following non-convex constrained problem:
\begin{equation}
\begin{aligned}
& \underset{Net}{\text{maximize}}
& & Acc(Net) \\
& \text{subject to}
& & Res_j(Net) \leq Bud_j, \; j = 1, \ldots, m,
\end{aligned}
\end{equation}
where $Net$ is a simplified network from the initial pretrained network, $Acc(\cdot)$ computes the accuracy, $Res_j(\cdot)$ evaluates the direct metric for resource consumption of the $j^{th}$ resource, and $Bud_j$ is the budget of the $j^{th}$ resource and the constraint on the optimization. The resource can be latency, energy, memory footprint, etc., or a combination of these metrics. 

Based on an idea similar to progressive barrier methods~\cite{siam2009-audet-progressive-barrier}, NetAdapt breaks this problem into the following series of easier problems and solves it iteratively:
\begin{equation}
\begin{aligned}
& \underset{Net_i}{\text{maximize}}
& & Acc(Net_i) \\
& \text{subject to}
& & Res_j(Net_i) \leq Res_j(Net_{i-1})-\Delta R_{i,j}, \; j = 1, \ldots, m,
\label{eq:optimization}
\end{aligned}
\end{equation}
where $Net_{i}$ is the network generated by the $i^{th}$ iteration, and $Net_0$ is the initial pretrained network. As the number of iterations increases, the constraints (i.e., current resource budget $Res_j(Net_{i-1})-\Delta R_{i,j}$) gradually become tighter.  $\Delta R_{i,j}$, which is larger than zero, indicates how much the constraint tightens for the $j^{th}$ resource in the $i^{th}$ iteration and can vary from iteration to iteration.  This is referred to as ``resource reduction schedule", which is similar to the concept of learning rate schedule. The algorithm terminates when $Res_j(Net_{i-1})-\Delta R_{i,j}$ is equal to or smaller than $Bud_j$ for every resource type. It outputs the final adapted network and can also generate a sequence of simplified networks (i.e., the highest accuracy network from each iteration $Net_{1},...,Net_{i}$) to provide the efficient frontier of accuracy and resource consumption trade-offs.

\subsection{Algorithm Overview}

\begin{algorithm}[!t]
  \caption{NetAdapt}
  \label{cod:netadapt}
  \SetNoFillComment
  \KwIn{Pretrained Network: $Net_0$ (with $K$ CONV and FC layers), Resource Budget: $Bud$, Resource Reduction Schedule: $\Delta R_{i}$}
  \KwOut{Adapted Network Meeting the Resource Budget: $\hat{Net}$}
  i = 0\;
  $Res_i$ = TakeEmpiricalMeasurement($Net_i$)\;
  \While{$Res_i$ $>$ Bud}
  {
    Con = $Res_i$ - $\Delta R_i$\;
    \For{k from 1 to K}
    {
      \tcc{TakeEmpiricalMeasurement is also called inside ChooseNumFilters for choosing the correct number of filters that satisfies the constraint (i.e., current budget).}
      $N\_Filt_k$, $Res\_Simp_k$ = ChooseNumFilters($Net_i$, k, Con)\;
      $Net\_Simp_k$ = ChooseWhichFilters($Net_i$, k, $N\_Filt_k$)\;
      $Net\_Simp_k$ = ShortTermFineTune($Net\_Simp_k$)\;
    }
    $Net_{i+1}$, $Res_{i+1}$ = PickHighestAccuracy($Net\_Simp_:$, $Res\_Simp_:$)\;
    i = i + 1\;
  }
  $\hat{Net}$ = LongTermFineTune($Net_{i}$)\;
  return $\hat{Net}$\;
\end{algorithm}

\begin{figure}[!t]
    \centering
    \includegraphics[width=0.7\textwidth]{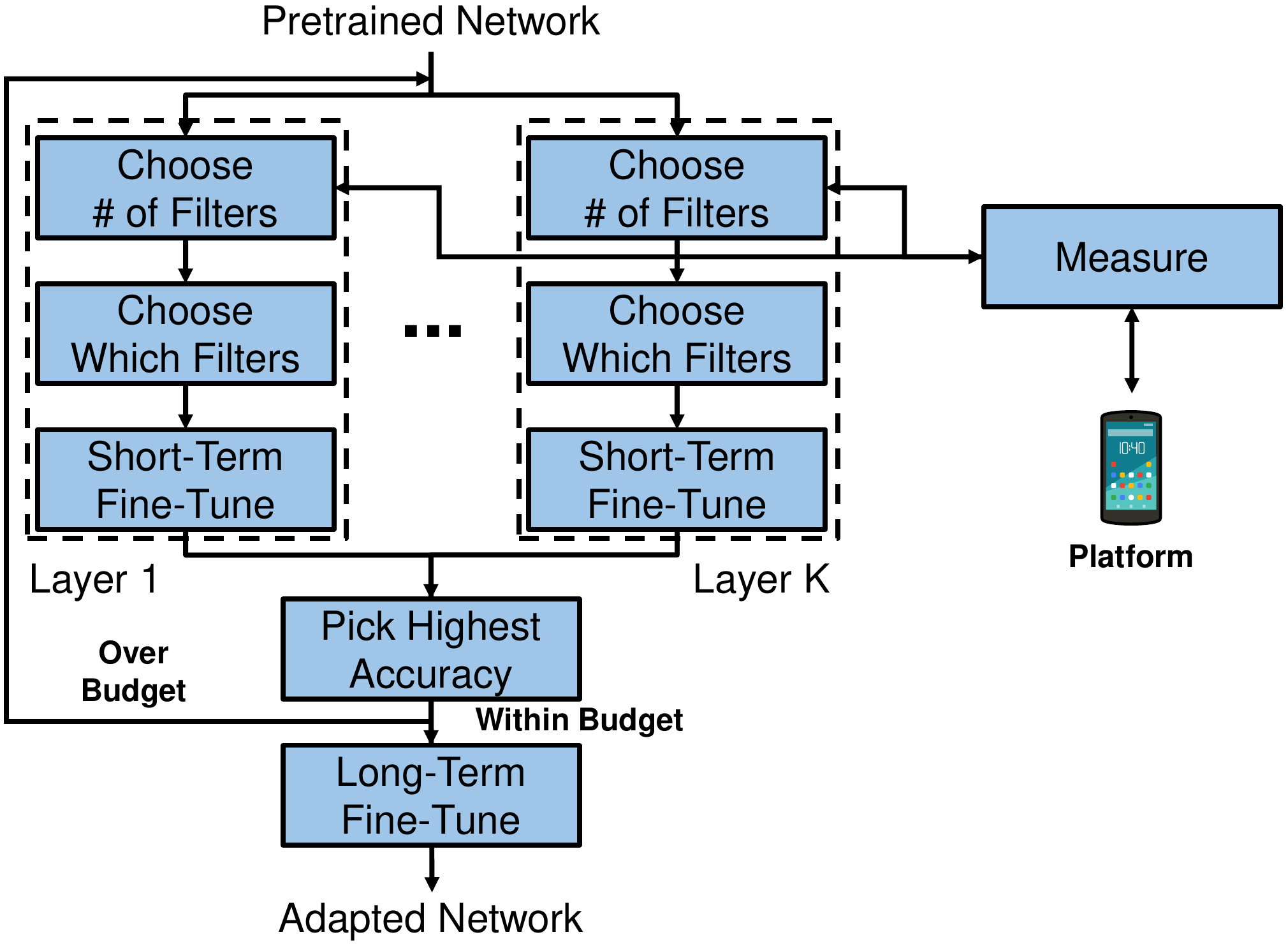}
    \caption{This figure visualizes the algorithm flow of NetAdapt. At each iteration, NetAdapt decreases the resource consumption by simplifying (i.e., removing filters from) one layer. In order to maximize accuracy, it tries to simplify each layer individually and picks the simplified network that has the highest accuracy. Once the target budget is met, the chosen network is then fine-tuned again until convergence.}
    \label{fig:algorithm_flow}
\end{figure}

For simplicity, we assume that we only need to meet the budget of one resource, specifically latency. One method to reduce the latency is to remove filters from the convolutional (CONV) or fully-connected (FC) layers. While there are other ways to reduce latency, we will use this approach to demonstrate NetAdapt.

The NetAdapt algorithm is detailed in pseudo code in Algorithm~\ref{cod:netadapt} and in Fig.~\ref{fig:algorithm_flow}. 
Each iteration solves Eq.~\ref{eq:optimization} by reducing the number of filters in a \emph{single} CONV or FC layer (the \textbf{Choose \# of Filters} and \textbf{Choose Which Filters} blocks in Fig.~\ref{fig:algorithm_flow}). The number of filters to remove from a layer is guided by empirical measurements. NetAdapt removes entire filters instead of individual weights because most platforms can take advantage of removing entire filters, and this strategy allows reducing both filters and feature maps, which play an important role in resource consumption~\cite{cvpr2017-yang-energy-aware-pruning}.  The simplified network is then fine-tuned for a short length of time in order to restore some accuracy (the \textbf{Short-Term Fine-Tune} block). 

In each iteration, the previous three steps (highlighted in bold) are applied on each of the CONV or FC layers individually\footnote{The algorithm can also be applied to a group of multiple layers as a single unit (instead of a single layer). For example, in ResNet~\cite{cvpr2016-he-resnet}, we can treat a residual block as a single unit to speed up the adaptation process.}. As a result, NetAdapt generates K (i.e., the number of CONV and FC layers) network proposals in one iteration, each of which has a single layer modified from the previous iteration. The network proposal with the highest accuracy is carried over to the next iteration (the \textbf{Pick Highest Accuracy} block). Finally, once the target budget is met, the chosen network is fine-tuned again until convergence (the \textbf{Long-Term Fine-Tune} block). 

\subsection{Algorithm Details}
\label{ssec:algorithm_details}
This section describes the key blocks in the \emph{NetAdapt} algorithm (Fig.~\ref{fig:algorithm_flow}).

\textbf{Choose Number of Filters}
This step focuses on determining \emph{how many} filters to preserve in a specific layer based on empirical measurements. NetAdapt gradually reduces the number of filters in the target layer and measures the resource consumption of each of the simplified networks. The maximum number of filters that can satisfy the current resource constraint will be chosen. Note that when some filters are removed from a layer, the associated channels in the following layers should also be removed. Therefore, the change in the resource consumption of other layers needs to be factored in.

\textbf{Choose Which Filters}
This step chooses \emph{which} filters to preserve based on the architecture from the previous step. There are many methods proposed in the literature, and we choose the magnitude-based method to keep the algorithm simple. In this work, the $N$ filters that have the largest $\ell$2-norm magnitude will be kept, where $N$ is the number of filters determined by the previous step. More complex methods can be adopted to increase the accuracy, such as removing the filters based on their joint influence on the feature maps~\cite{cvpr2017-yang-energy-aware-pruning}.

\textbf{Short-/Long-Term Fine-Tune}
Both the short-term fine-tune and long-term fine-tune steps in NetAdapt involve network-wise end-to-end fine-tuning. Short-term fine-tune has fewer iterations than long-term fine-tune.

At each iteration of the algorithm, we fine-tune the simplified networks with a relatively smaller number of iterations (i.e., short-term) to regain accuracy, in parallel or in sequence. This step is especially important while adapting small networks with a large resource reduction because otherwise the accuracy will drop to zero, which can  cause the algorithm to choose the wrong network proposal.

As the algorithm proceeds, the network is continuously trained but does not converge. Once the final adapted network is obtained, we fine-tune the network with more iterations until convergence (i.e., long-term) as the final step.

\subsection{Fast Resource Consumption Estimation}
\label{subsec:latency_approx}

\begin{figure}[!t]
    \centering
    \includegraphics[width=0.7\textwidth]{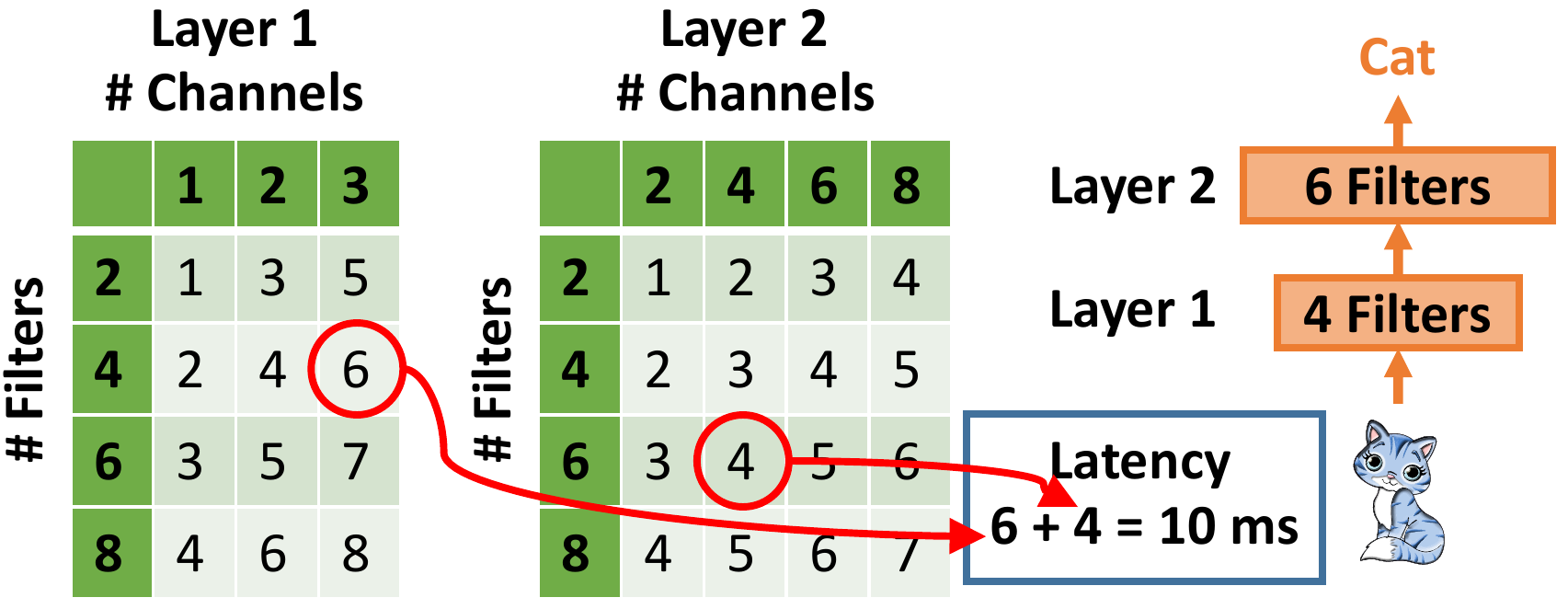}
    \caption{This figure illustrates how layer-wise look-up tables are used for fast resource consumption estimation.}
    \label{fig:lookup_table}
\end{figure}

\begin{figure}[!t]
    \centering
    \includegraphics[width=0.5\textwidth]{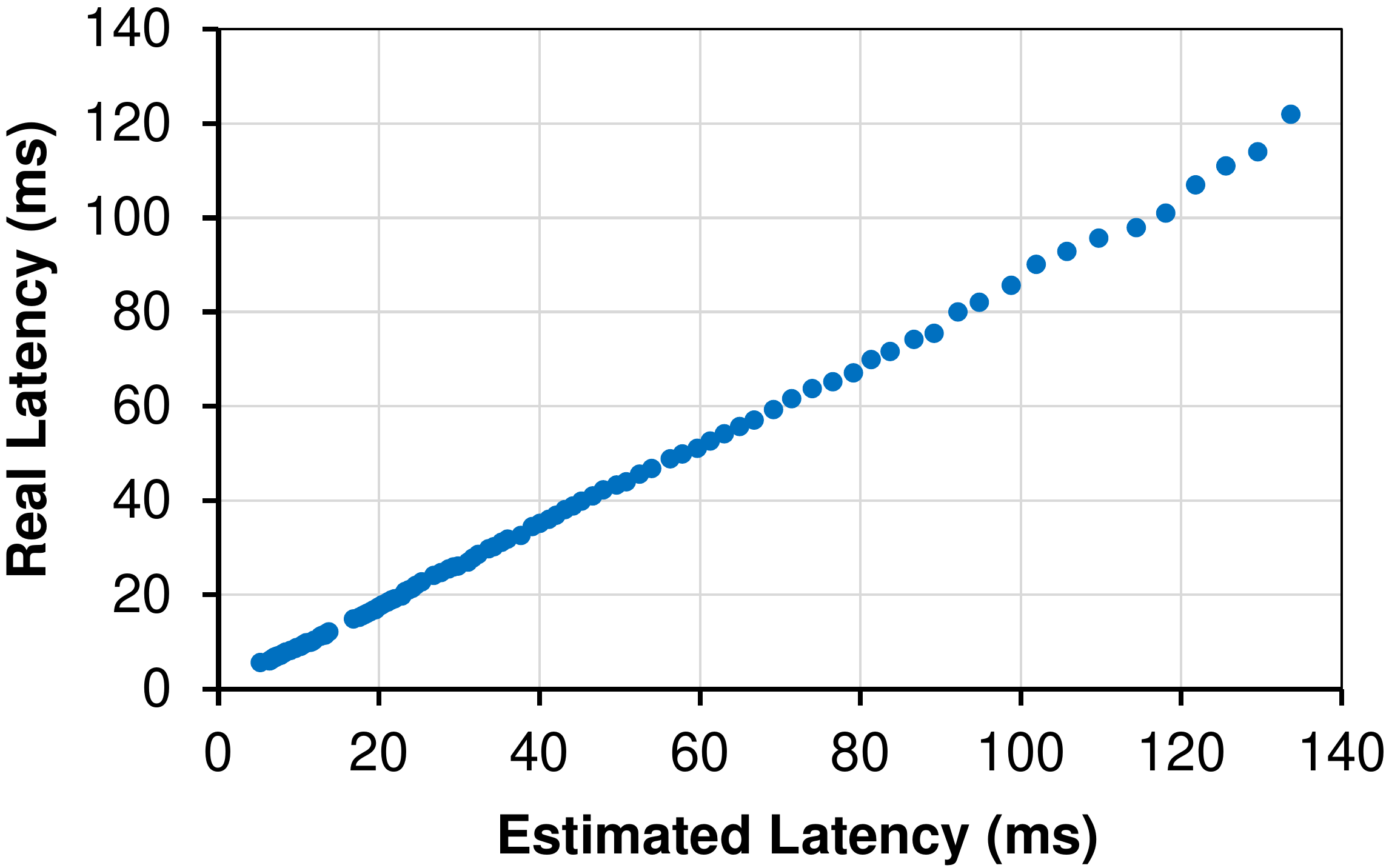}
    \caption{The comparison between the estimated latency (using layer-wise look-up tables) and the real latency on a single large core of Google Pixel 1 CPU while adapting the 100\% MobileNetV1 with the input resolution of 224~\cite{howard2017mobilenets}.}
    \label{fig:estimated_vs_real_latency}
\end{figure}

As mentioned in Sec.~\ref{ssec:algorithm_details}, NetAdapt uses empirical measurements to determine the number of filters to keep in a layer given the resource constraint. In theory, we can measure the resource consumption of each of the simplified networks on the fly during adaptation. However, taking measurements can be slow and difficult to parallelize due to the limited number of available devices. Therefore, it may be prohibitively expensive and become the computation bottleneck.

We solve this problem by building layer-wise look-up tables with pre-measured resource consumption of each layer. When executing the algorithm, we look up the table of each layer, and sum up the layer-wise measurements to estimate the network-wise resource consumption, which is illustrated in Fig.~\ref{fig:lookup_table}. The reason for not using a network-wise table is that the size of the table will grow exponentially with the number of layers, which makes it intractable for deep networks. Moreover, layers with the same shape and feature map size only need to be measured once, which is common for modern deep networks.

Fig.~\ref{fig:estimated_vs_real_latency} compares the estimated latency (the sum of layer-wise latency from the layer-wise look-up tables) and the real latency on a single large core of Google Pixel 1 CPU while adapting the 100\% MobileNetV1 with the input resolution of 224~\cite{howard2017mobilenets}. The real and estimated latency numbers are highly correlated, and the difference between them is sufficiently small to be used by NetAdapt.

\section{Experiment Results}
\label{sec:experiment_results}

In this section, we apply the proposed NetAdapt algorithm to MobileNets~\cite{howard2017mobilenets,cvpr2018-sandler-mobilenet-v2}, which are designed for mobile applications, and experiment on the ImageNet dataset~\cite{imagenet}. We did not apply NetAdapt on larger networks like ResNet~\cite{cvpr2016-he-resnet} and VGG~\cite{iclr2015-simonyan-vgg} because networks become more difficult to simplify as they become smaller; these networks are also seldom deployed on mobile platforms. We benchmark NetAdapt against three state-of-the-art network simplification methods:
\begin{itemize}
  \item \textbf{Multipliers}~\cite{howard2017mobilenets} are simple but effective methods for simplifying networks. Two commonly used multipliers are the width multiplier and the resolution multiplier; they can also be used together. Width multiplier scales the number of filters by a percentage across all convolutional (CONV) and fully-connected (FC) layers, and resolution multiplier scales the resolution of the input image. We use the notation ``50\% MobileNetV1 (128)" to denote applying a width multiplier of 50\% on MobileNetV1 with the input image resolution of 128.
  \item \textbf{MorphNet}~\cite{gordon2018morphnet} is an automatic network simplification algorithm based on sparsifying regularization.
  \item \textbf{ADC}~\cite{he2018adc} is an automatic network simplification algorithm based on reinforcement learning.
\end{itemize}

We will show the performance of NetAdapt on the small MobileNetV1 (50\% MobileNetV1 (128)) to demonstrate the effectiveness of NetAdapt on real-time-class networks, which are much more difficult to simplify than larger networks. To show the generality of NetAdapt, we will also measure its performance on the large MobileNetV1 (100\% MobileNetV1 (224)) across different platforms. Lastly, we adapt the large MobileNetV2 (100\% MobileNetV2 (224)) to push the frontier of efficient networks.

\subsection{Detailed Settings for MobileNetV1 Experiments}
We perform most of the experiments and study on MobileNetV1 and detail the settings in this section.

\textbf{NetAdapt Configuration} MobileNetV1~\cite{howard2017mobilenets} is based on depthwise separable convolutions, which factorize a $m \times m$ standard convolution layer into a $m \times m$ depthwise layer and a $1 \times 1$ standard convolution layer called a pointwise layer. In the experiments, we adapt each depthwise layer with the corresponding pointwise layer and choose the filters to keep based on the pointwise layer. When adapting the small MobileNetV1 (50\% MobileNetV1 (128)), the latency reduction ($\Delta R_{i,j}$ in Eq.~\ref{eq:optimization}) starts at 0.5 and decays at the rate of 0.96 per iteration. When adapting other networks, we use the same decay rate but scale the initial latency reduction proportional to the latency of the initial pretrained network.

\textbf{Network Training} We preserve ten thousand images from the training set, ten images per class, as the holdout set. The new training set without the holdout images is used to perform short-term fine-tuning, and the holdout set is used to pick the highest accuracy network out of the simplified networks at each iteration. The whole training set is used for the long-term fine-tuning, which is performed once in the last step of NetAdapt.

Because the training configuration can have a large impact on the accuracy, we apply the same training configuration to all the networks unless otherwise stated to have a fairer comparison. We adopt the same training configuration as MorphNet~\cite{gordon2018morphnet} (except that the batch size is 128 instead of 96). The learning rate for the long-term fine-tuning is 0.045 and that for the short-term fine-tuning is 0.0045. This configuration improves ADC network's top-1 accuracy by 0.3\% and almost all multiplier networks' top-1 accuracy by up to 3.8\%, except for one data point, whose accuracy is reduced by 0.2\%. We use these numbers in the following analysis. Moreover, all accuracy numbers are reported on the validation set to show the true performance.

\textbf{Mobile Inference and Latency Measurement} We use Google's TensorFlow Lite engine~\cite{tflite} for inference on a mobile CPU and Qualcomm's Snapdragon Neural Processing Engine (SNPE) for inference on a mobile GPU. For experiments on mobile CPUs, the latency is measured on a single large core of Google Pixel 1 phone. For experiments on mobile GPUs, the latency is measured on the mobile GPU of Samsung Galaxy S8 with SNPE's benchmarking tool. For each latency number, we report the median of 11 latency measurements.

\subsection{Comparison with Benchmark Algorithms}
\label{subsec:comparison}
\subsubsection{Adapting Small MobileNetV1 on a Mobile CPU}

\begin{figure}[!t]
    \centering
    \includegraphics[width=0.85\textwidth]{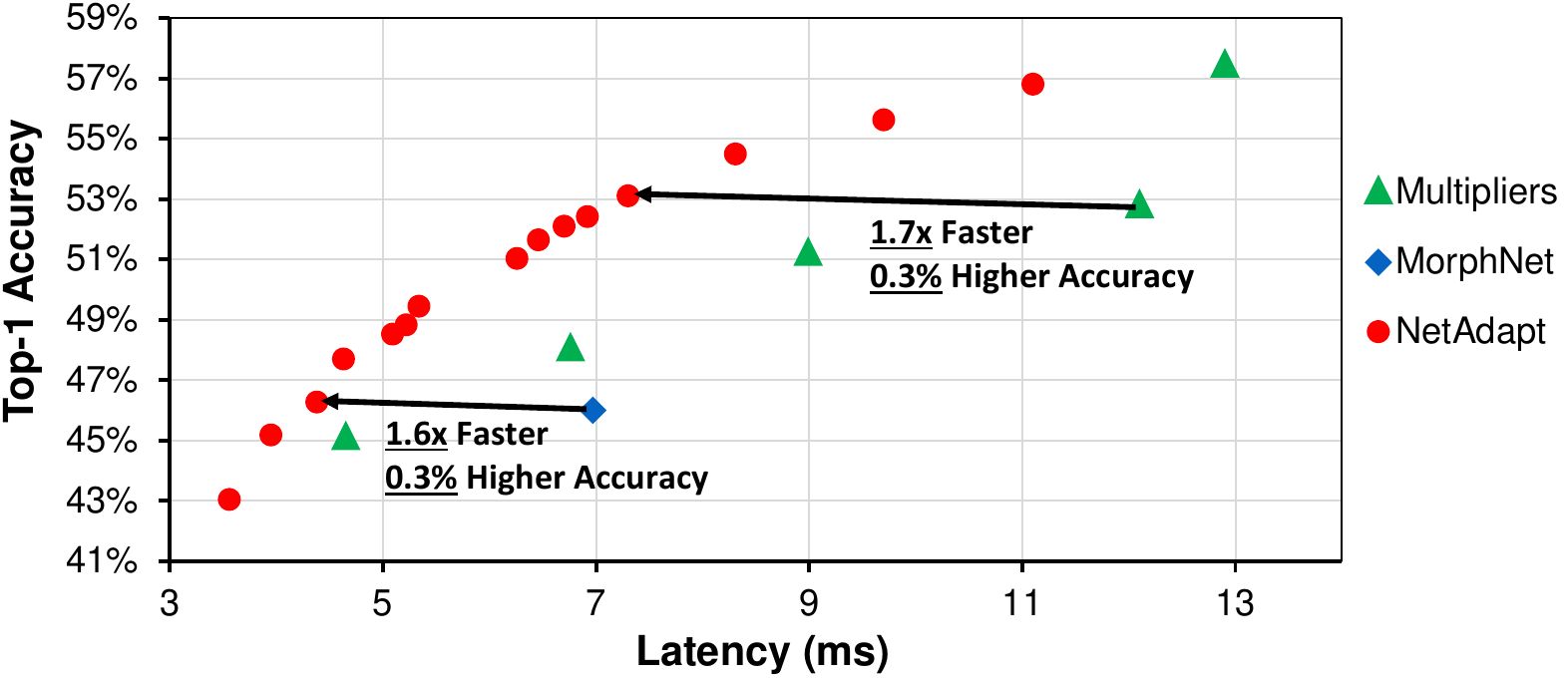}
    \caption[Caption for Fig]{The figure compares NetAdapt (adapting the small MobileNetV1) with the multipliers~\cite{howard2017mobilenets} and MorphNet~\cite{gordon2018morphnet} on a mobile CPU of Google Pixel 1.}
    \label{fig:mobilenet_small_cpu}
\end{figure}

In this experiment, we apply NetAdapt to adapt the small MobileNetV1 (50\% MobileNetV1 (128)) to a mobile CPU. It is one of the most compact networks and achieves real-time performance. It is more challenging to simplify than other larger networks (include the large MobileNetV1). The results are summarized and compared with the multipliers~\cite{howard2017mobilenets} and MorphNet~\cite{gordon2018morphnet} in Fig.~\ref{fig:mobilenet_small_cpu}. We observe that NetAdapt outperforms the multipliers by up to 1.7$\times$ faster with the same or higher accuracy. For MorphNet, NetAdapt's result is 1.6$\times$ faster with 0.3\% higher accuracy.

\subsubsection{Adapting Large MobileNetV1 on a Mobile CPU}
\label{subsubsec:mobilenet_large_cpu}

\begin{figure}[!t]
    \centering
    \includegraphics[width=0.85\textwidth]{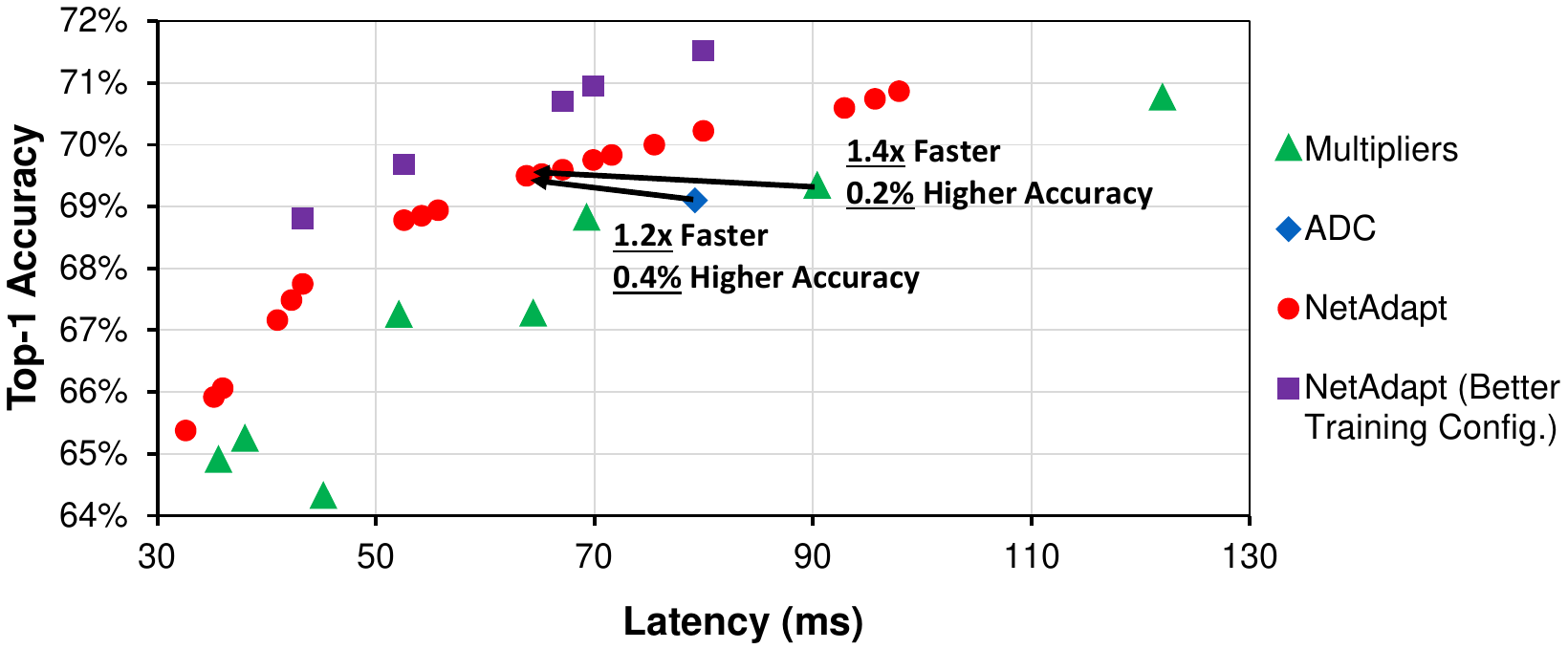}
    \caption{The figure compares NetAdapt (adapting the large MobileNetV1) with the multipliers~\cite{howard2017mobilenets} and ADC~\cite{he2018adc} on a mobile CPU of Google Pixel 1. Moreover, the accuracy of the adapted networks can be further increased by up to 1.3\% through using a better training configuration (simply adding dropout and label smoothing).}
    \label{fig:mobilenet_large_cpu}
\end{figure}

In this experiment, we apply NetAdapt to adapt the large MobileNetV1 (100\% MobileNetV1 (224)) on a mobile CPU. It is the largest MobileNetV1 and achieves the highest accuracy. Because its latency is approximately 8$\times$ higher than that of the small MobileNetV1, we scale the initial latency reduction by 8$\times$. The results are shown and compared with the multipliers~\cite{howard2017mobilenets} and ADC~\cite{he2018adc} in Fig.~\ref{fig:mobilenet_large_cpu}. NetAdapt achieves higher accuracy than the multipliers and ADC while increasing the speed by 1.4$\times$ and 1.2$\times$, respectively.

While the training configuration is kept the same when comparing to the benchmark algorithms discussed above, we also show in Fig.~\ref{fig:mobilenet_large_cpu} that the accuracy of the networks adapted using NetAdapt can be further improved with a better training configuration. After simply adding dropout and label smoothing, the accuracy can be increased by 1.3\%. Further tuning the training configuration for each adapted network can give higher accuracy numbers, but it is not the focus of this paper.

\subsubsection{Adapting Large MobileNetV1 on a Mobile GPU}

\begin{figure}[!t]
    \centering
    \includegraphics[width=0.85\textwidth]{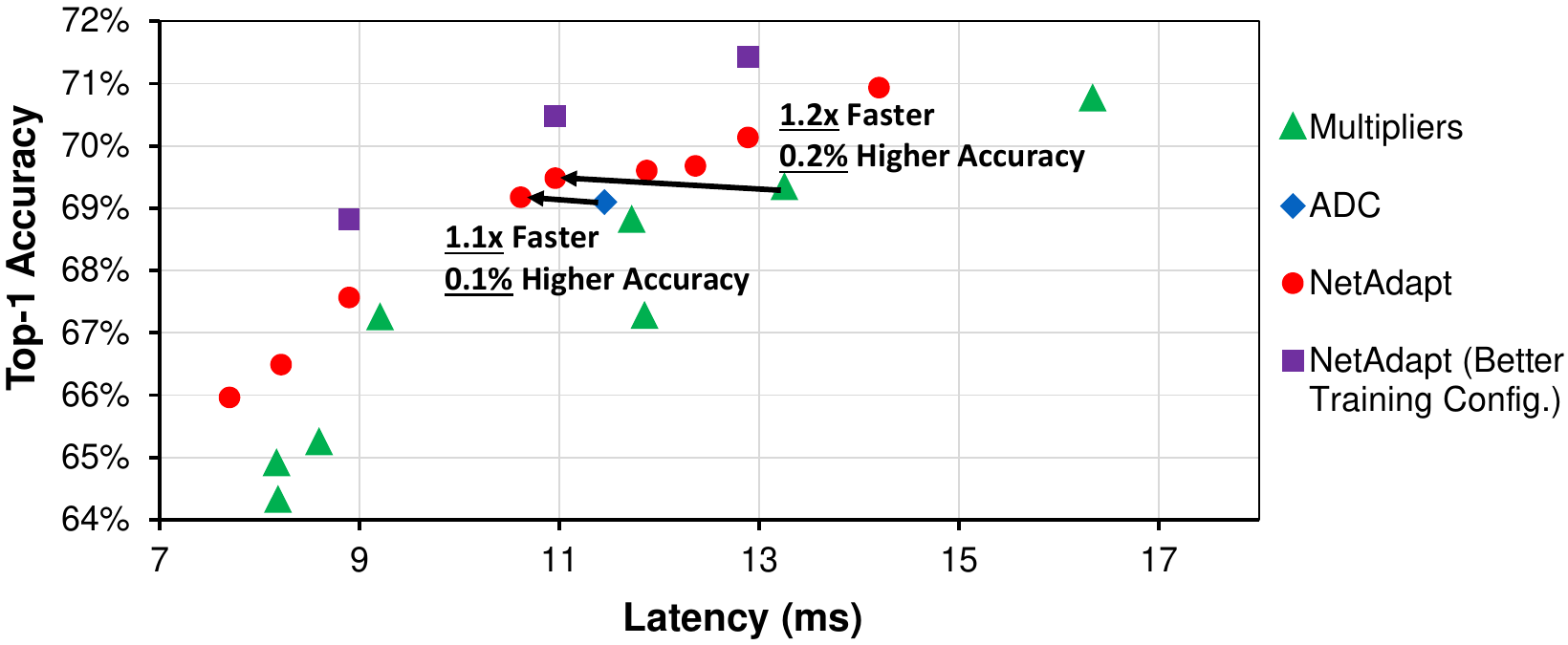}
    \caption{This figure compares NetAdapt (adapting the large MobileNetV1) with the multipliers~\cite{howard2017mobilenets} and ADC~\cite{he2018adc} on a mobile GPU of Samsung Galaxy S8. Moreover, the accuracy of the adapted networks can be further increased by up to 1.3\% through using a better training configuration (simply adding dropout and label smoothing).}
    \label{fig:mobilenet_large_gpu}
\end{figure}

In this experiment, we apply NetAdapt to adapt the large MobileNetV1 on a mobile GPU to show the generality of NetAdapt. Fig.~\ref{fig:mobilenet_large_gpu} shows that NetAdapt outperforms other benchmark algorithms by up to 1.2$\times$ speed-up with higher accuracy. Due to the limitation of the SNPE tool, the layerwise latency breakdown only considers the computation time and does not include the latency of other operations, such as feature map movement, which can be expensive~\cite{cvpr2017-yang-energy-aware-pruning}. This affects the precision of the look-up tables used for this experiment. Moreover, we observe that there is an approximate 6.2ms (38\% of the latency of the network before applying NetAdapt) non-reducible latency. These factors cause a smaller improvement on the mobile GPU compared with the experiments on the mobile CPU. Moreover, when the better training configuration is applied as previously described, the accuracy can be further increased by 1.3\%.

\subsection{Ablation Studies}

\subsubsection{Impact of Direct Metrics}
\label{subsec:results_on_macs}

\begin{table}[!t]
\centering
\resizebox{\textwidth}{!}{
    \begin{tabular}{c | C C | C C | C C} 
     \hline
     \multicolumn{1}{c}{\textbf{Network}} & \multicolumn{2}{|c|}{\textbf{Top-1 Accuracy (\%)}} & \multicolumn{2}{c}{\textbf{\# of MACs ($\times 10^6$)}} & \multicolumn{2}{|c}{\textbf{Latency (ms)}} \\
     \hline
     25\% MobileNetV1 (128)~\cite{howard2017mobilenets} & 45.1 & (+0) & 13.6 & (100\%) & 4.65 & (100\%) \\ 
     MorphNet~\cite{gordon2018morphnet} & 46.0 & (+0.9) & 15.0 & (110\%) & 6.52 & (140\%) \\
     NetAdapt & 46.3 & (+1.2) & 11.0 & (81\%) & 6.01 & (129\%) \\
     \hline
     75\% MobileNetV1 (224)~\cite{howard2017mobilenets} & 68.8 & (+0)   & 325.4 & (100\%) & 69.3 & (100\%) \\
     ADC~\cite{he2018adc} & 69.1 & (+0.3)   & 304.2 & (93\%)  & 79.2 & (114\%) \\
     NetAdapt             & 69.1 & (+0.3) & 284.3 & (87\%)  & 74.9 & (108\%) \\
     \hline
    \end{tabular}
}
\caption{The comparison between NetAdapt (adapting the small or large MobileNetV1) and the three benchmark algorithms on image classification when targeting the number of MACs. The latency numbers are measured on a mobile CPU of Google Pixel 1. We roughly match their accuracy and compare their latency.}
\label{table:mobilenet_cpu_benchmark_macs}
\end{table}

In this experiment, we use the indirect metric (i.e., the number of MACs) instead of the direct metric (i.e., the latency) to guide NetAdapt to investigate the importance of using direct metrics. When computing the number of MACs, we only consider the CONV and FC layers because batch normalization layers can be folded into the corresponding CONV layers, and the other layers are negligibly small. Table~\ref{table:mobilenet_cpu_benchmark_macs} shows that NetAdapt outperforms the benchmark algorithms with lower numbers of MACs and higher accuracy. This demonstrates the effectiveness of NetAdapt. However, we also observe that the network with lower numbers of MACs may not necessarily be faster. This shows the necessity of incorporating direct measurements into the optimization flow.

\subsubsection{Impact of Short-Term Fine-Tuning}
\label{subsubsec:exp_short_term_finetuning}

\begin{figure}[!t]
  \begin{minipage}[t]{0.48\textwidth}
    \centering
    \includegraphics[width=1.0\textwidth]{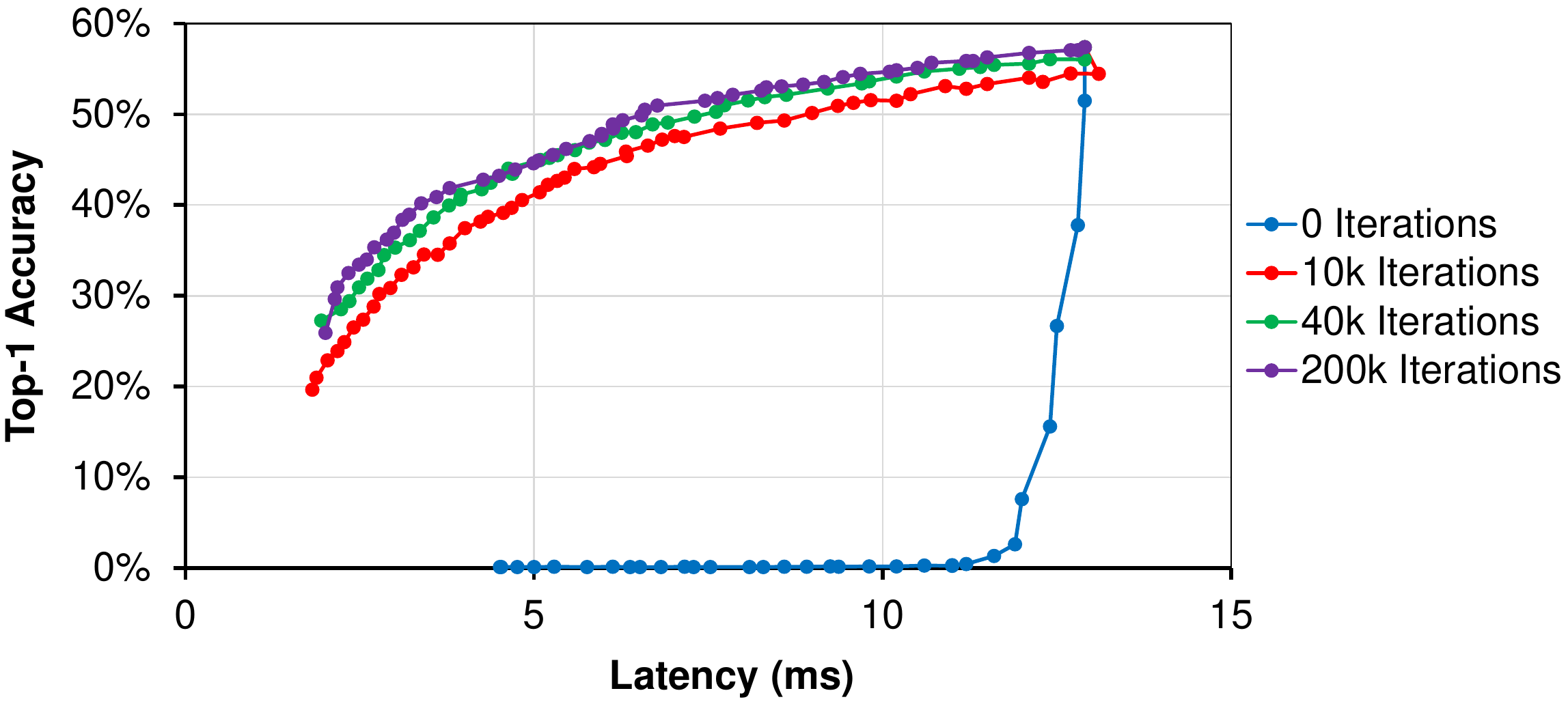}
    \caption{The accuracy of different short-term fine-tuning iterations when adapting the small MobileNetV1 (without long-term fine-tuning) on a mobile CPU of Google Pixel 1. Zero iterations means no short-term fine-tuning.}
    \label{fig:short_term_finetuning}
  \end{minipage}%
  \hfill
  \begin{minipage}[t]{0.48\textwidth}
    \centering
    \includegraphics[width=1.0\textwidth]{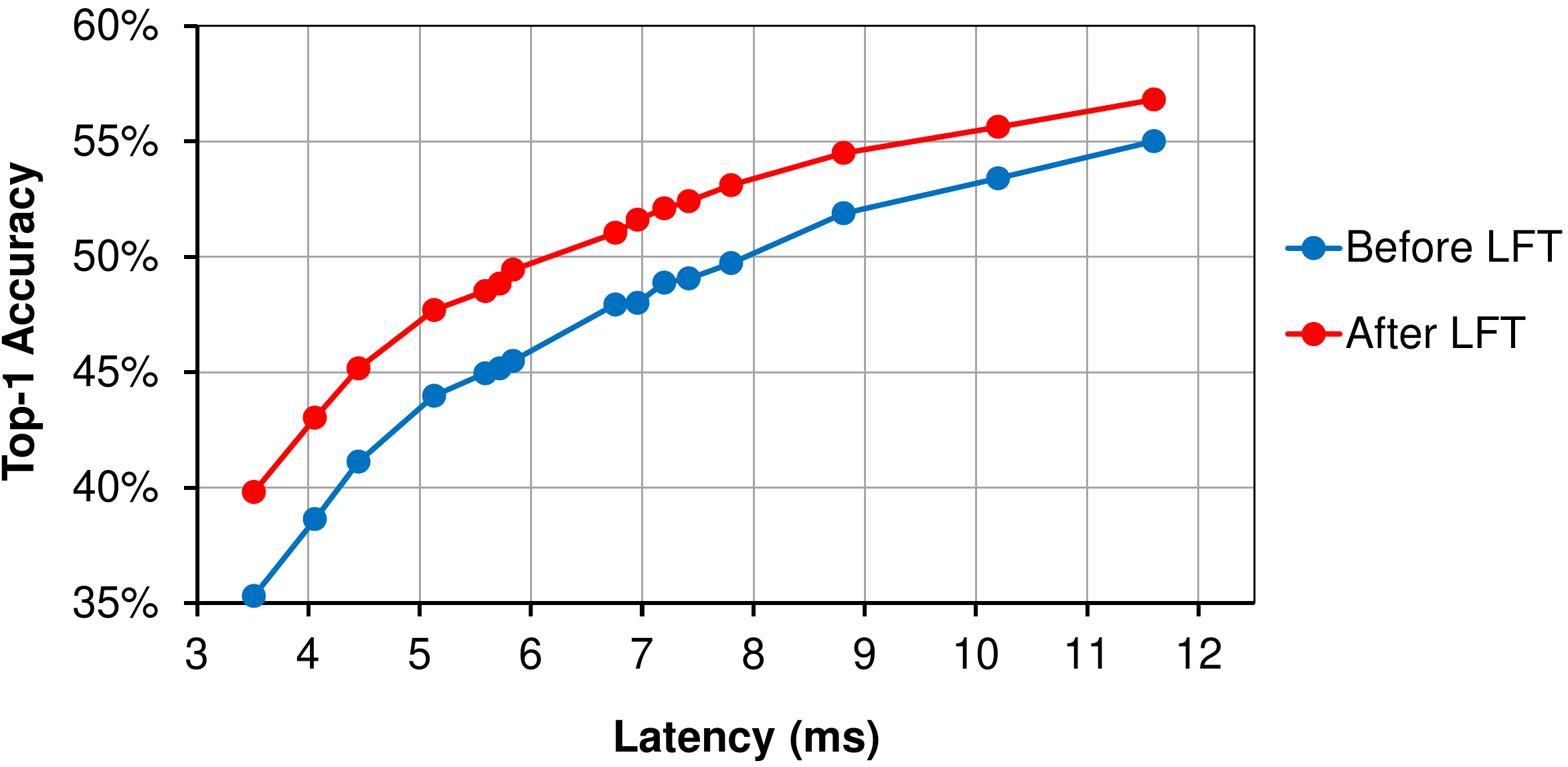}
    \caption{The comparison between before and after long-term fine-tuning when adapting the small MobileNetV1 on a mobile CPU of Google Pixel 1. Although the short-term fine-tuning preserves the accuracy well, the long-term fine-tuning  gives the extra 3.4\% on average (from 1.8\% to 4.5\%).}
    \label{fig:long_term_finetuning}
  \end{minipage}
\end{figure}

Fig.~\ref{fig:short_term_finetuning} shows the accuracy of adapting the small MobileNetV1 with different short-term fine-tuning iterations (without long-term fine-tuning). The accuracy rapidly drops to nearly zero if no short-term fine-tuning is performed (i.e., zero iterations). In this low accuracy region, the algorithm picks the best network proposal solely based on noise and hence gives poor performance. After fine-tuning a network for a short amount of time (ten thousand iterations), the accuracy is always kept above 20\%, which allows the algorithm to make a better decision. Although further increasing the number of iterations improves the accuracy, we find that using forty thousand iterations leads to a good accuracy versus speed trade-off for the small MobileNetV1.

\subsubsection{Impact of Long-Term Fine-Tuning}

Fig.~\ref{fig:long_term_finetuning} illustrates the importance of performing the long-term fine-tuning. Although the short-term fine-tuning preserves the accuracy well, the long-term fine-tuning can still increase the accuracy by up to another 4.5\% or 3.4\% on average. Since the short-term fine-tuning has a short training time, the training is terminated far before convergence. Therefore, it is not surprising that the final long-term fine-tuning can further increase the accuracy.

\subsubsection{Impact of Resource Reduction Schedules}
\label{subsec:resource_reduction_scheduling}

\begin{table}[!t]
\centering
\resizebox{\textwidth}{!}{
    \begin{tabular}{c | c | c | c | c} 
     \hline
     Initialization (ms) & Decay Rate & \# of Total Iterations & Top-1 Accuracy (\%) & Latency (ms) \\
     \hline
     0.5 & 0.96 & 28 & 47.7 & 4.63 \\ 
     0.5 & 1.0 & 20 & 47.4 & 4.71 \\
     0.8 & 0.95 & 20 & 46.7 & 4.65 \\
     \hline
    \end{tabular}
}
\caption{The influence of resource reduction schedules.}
\label{table:resource_reduction_scheduling}
\end{table}

Table~\ref{table:resource_reduction_scheduling} shows the impact of using three different resource reduction schedules, which are defined in Sec.~\ref{subsec:problem_formulation}. Empirically, using a larger resource reduction at each iteration increases the adaptation speed (i.e., reducing the total number of adaptation iterations) at the cost of accuracy. With the same number of total iterations, the result suggests that a smaller initial resource reduction with a slower decay is preferable.

\subsection{Analysis of Adapted Network Architecture}

\begin{figure}[!t]
    \centering
    \includegraphics[width=0.6\textwidth]{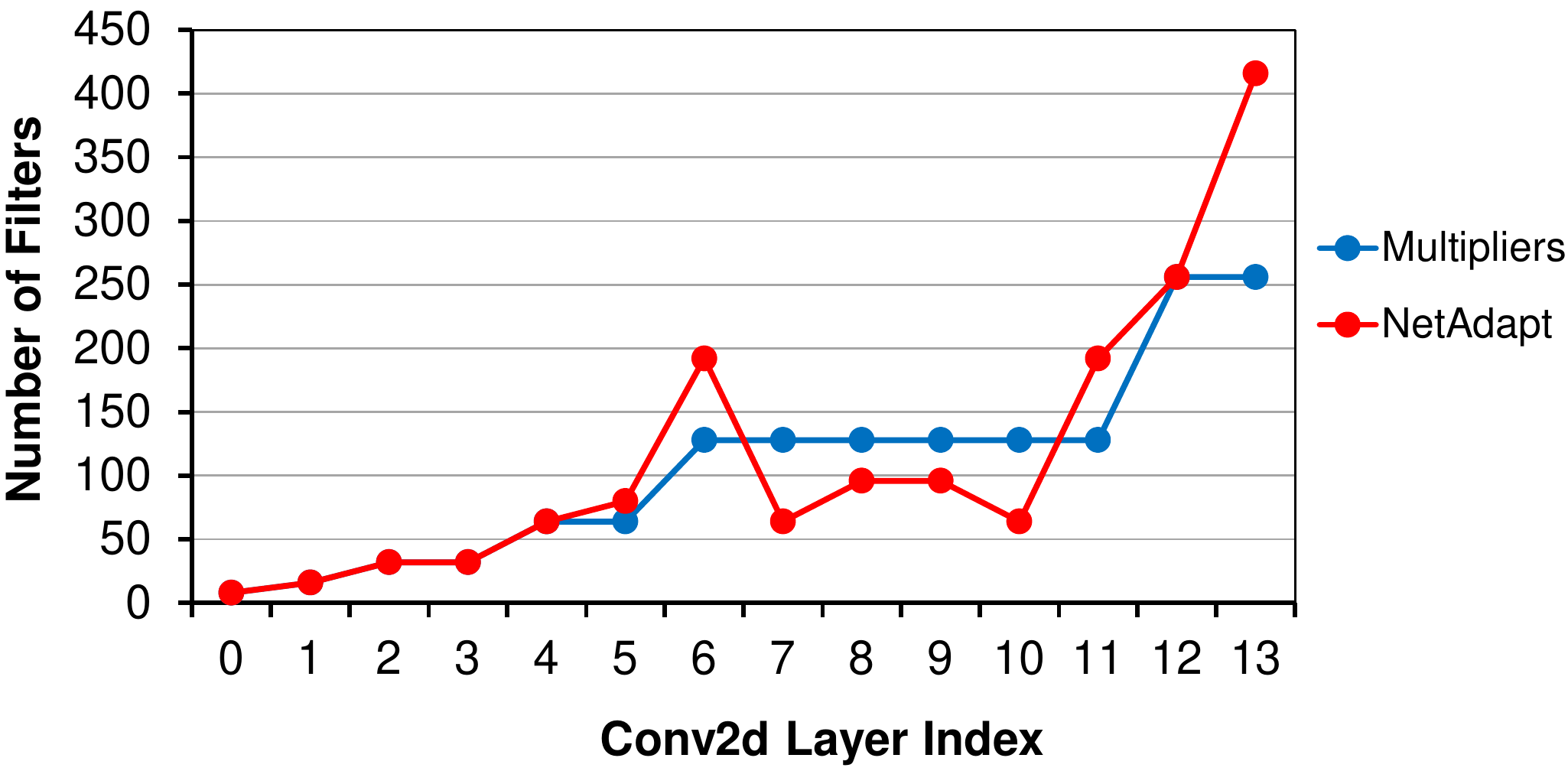}
    \caption{NetAdapt and the multipliers generate different simplified networks when adapting the small MobileNetV1 to match the latency of 25\% MobileNetV1 (128).}
    \label{fig:shape_comparison}
\end{figure}

The network architectures of the adapted small MobileNetV1 by using NetAdapt and the multipliers are shown and compared in Fig.~\ref{fig:shape_comparison}. Both of them have similar latency as 25\% MobileNetV1 (128). There are two interesting observations. 

First, NetAdapt removes more filters in layers $7$ to $10$, but fewer in layer $6$. Since the feature map resolution is reduced in layer $6$ but not in layers $7$ to $10$, we hypothesize that when the feature map resolution is reduced, more filters are needed to avoid creating an information bottleneck.

The second observation is that NetAdapt keeps more filters in layer $13$ (i.e. the last CONV layer). One possible explanation is that the ImageNet dataset contains one thousand classes, so more feature maps are needed by the last FC layer to do the correct classification.

\subsection {Adapting Large MobileNetV2 on a Mobile CPU}

\begin{table}[!t]
\centering
\begin{tabular}{c | C C | C C} 
 \hline
 \multicolumn{1}{c}{\textbf{Network}} & \multicolumn{2}{|c|}{\textbf{Top-1 Accuracy (\%)}} & \multicolumn{2}{c}{\textbf{Latency (ms)}} \\
 \hline
 75\% MobileNetV2 (224)~\cite{cvpr2018-sandler-mobilenet-v2} & 69.8 & (+0) & 64.5 & (100\%) \\ 
 NetAdapt (Similar Latency)  & 70.9 & (+1.1) & 63.6 & (99\%) \\
 NetAdapt (Similar Accuracy) & 70.2 & (+0.4) & 55.5 & (86\%) \\
 \hline
\end{tabular}
\caption{The comparison between NetAdapt (adapting the large MobileNetV2 (100\% MobileNetV2 (224))) and the multipliers~\cite{cvpr2018-sandler-mobilenet-v2} on a mobile CPU of Google Pixel 1. We compare the latency at similar accuracy and the accuracy at similar latency.}
\label{table:mobilenet_v2_large_cpu_highlight}
\end{table}

In this section, we show encouraging early results of applying NetAdapt to MobileNetV2 \cite{cvpr2018-sandler-mobilenet-v2}. MobileNetV2 introduces the inverted residual with linear bottleneck into MobileNetV1 and becomes more efficient. Because MobileNetV2 utilizes residual connections, we only adapt individual inner (expansion) layers or reduce all bottleneck layers of the same resolution in lockstep. The main differences between the MobileNetV1 and MobileNetV2 experiment settings are that each network proposal is short-term fine-tuned with ten thousand iterations, the initial latency reduction is 1ms, the latency reduction decay is 0.995, the batch size is 96, and dropout and label smoothing are used. NetAdapt achieves 1.1\% higher accuracy or 1.2$\times$ faster speed than the multipliers as shown in Table~\ref{table:mobilenet_v2_large_cpu_highlight}. 

\section{Conclusion}
\label{sec:conclusion}

In summary, we proposed an automated algorithm, called NetAdapt, to adapt a pretrained network to a mobile platform given a real resource budget. NetAdapt can incorporate direct metrics, such as latency and energy, into the optimization to maximize the adaptation performance based on the characteristics of the platform. By using empirical measurements, NetAdapt can be applied to any platform as long as we can measure the desired metrics, without any knowledge of the underlying implementation of the platform. We demonstrated empirically that the proposed algorithm can achieve better accuracy versus latency trade-off (by up to 1.7$\times$ faster with equal or higher accuracy) compared with other state-of-the-art network simplification algorithms. In this work, we aimed to highlight the importance of using direct metrics in the optimization of efficient networks; we hope that future research efforts will take direct metrics into account in order to further improve the performance of efficient networks.


\bibliographystyle{splncs04}
\bibliography{__references}

\end{document}